\documentclass[journal,transmag]{IEEEtran}
\ifCLASSINFOpdf
\else
\fi

\usepackage{graphicx}
\usepackage{multirow}
\usepackage[table,xcdraw]{xcolor}
\usepackage{rotating}
\usepackage{graphicx}
\usepackage{array}
\usepackage{stfloats}
\usepackage{verbatim}
\usepackage{amssymb}
\usepackage{graphicx}
\usepackage{ragged2e}
\usepackage{multirow}
\usepackage{booktabs}
\usepackage{hyperref}

\begin{document}
%
\title{A Spontaneous Driver Emotion Facial Expression (DEFE) Dataset for Intelligent Vehicles}


\author{Wenbo~Li,~\IEEEmembership{}
        Yaodong Cui,~\IEEEmembership{}
        Yintao Ma,~\IEEEmembership{}
        Xingxin Chen,~\IEEEmembership{}
        Guofa~Li,~\IEEEmembership{Member,~IEEE,}
        \\
        Gang~Guo,~\IEEEmembership{}
        and~Dongpu~Cao,~\IEEEmembership{Member,~IEEE}
\thanks{Wenbo Li and Gang Guo are with the School
of Automotive Engineering, Chongqing University, Chongqing, 400044, China (e-mail: liwenbocqu@foxmail.com, guogang@cqu.edu.cn).}
\thanks{Yaodong Cui, Yintao Ma, Xingxin Chen and Dongpu Cao are with the Department of Mechanical and Mechatronics Engineering, University of Waterloo, Waterloo, ON, N2L 3G1, Canada. Wenbo Li and Guofa Li are also with this affiliation. (e-mail: cuiyaodong@gmail.com, y354ma@uwaterloo.ca, xingxin.chen@uwaterloo.ca, dongpu.cao@uwaterloo.ca). }
\thanks{G. Li is with the Institute of Human Factors and Ergonomics, College of Mechatronics and Control Engineering, Shenzhen University, Shenzhen, Guangdong, 518060, China (e-mail: hanshan198@gmail.com).}

\thanks{}}

\markboth{}
{Shell \MakeLowercase{\textit{et al.}}: Bare Demo of IEEEtran.cls for IEEE Transactions on Magnetics Journals}
%



\IEEEtitleabstractindextext{%
\begin{abstract}
Abstract—In this paper, we introduce a new dataset, the driver emotion facial expression (DEFE) dataset, for driver spontaneous emotions analysis. The dataset includes facial expression recordings from 60 participants during driving. After watching a selected video-audio clip to elicit a specific emotion, each participant completed the driving tasks in the same driving scenario and rated their emotional responses during the driving processes from the aspects of dimensional emotion and discrete emotion. We also conducted classification experiments to recognize the scales of arousal, valence, dominance, as well as the emotion category and intensity to establish baseline results for the proposed dataset. Besides, this paper compared and discussed the differences in facial expressions between driving and non-driving scenarios. The results show that there were significant differences in AUs (Action Units) presence of facial expressions between driving and non-driving scenarios, indicating that human emotional expressions in driving scenarios were different from other life scenarios. Therefore, publishing a human emotion dataset specifically for the driver is necessary for traffic safety improvement. The proposed dataset will be publicly available so that researchers worldwide can use it to develop and examine their driver emotion analysis methods. To the best of our knowledge, this is currently the only public driver facial expression dataset.

\end{abstract}

}

\maketitle

\IEEEdisplaynontitleabstractindextext

%
\IEEEpeerreviewmaketitle

\section{Background and Related Work}
%
%
%
%

\IEEEPARstart{D}{river} emotion plays a vital role in driving because it affects driving safety and comfort. Among the 20-50 million non-fatal injuries and 1.24 million fatal road traffic accidents occurring every year worldwide \cite{organization_global_2015}, driver’s inability to control his emotions has been regarded as one of the critical factors degrading driving safety \cite{james_road_2000}\cite{li2020Influence}. The rapid development in intelligent vehicles also calls for an emerging demand in the integration of driver-automation interaction and collaboration to enhance driving comfort, where driver emotion is one of the critical states \cite{li2019detection}. Therefore, recognizing driver emotions is essential to improve driving safety and comfort of intelligent vehicles \cite{eyben2010emotion}.

\par To describe human emotion, psychological researchers have provided two methodologies to classify emotions, which are discrete emotions and dimensional emotions \cite{zeng2008survey}. Due to the discrete language words used by humans to describe emotions, discrete models are well-established and widely-accepted, such as the basic emotions of Ekman et al. \cite{ekman1987universals} and the emotion tree structure of Parrott \cite{parrott2001emotions}. Specifically, Ekman et al. categorized discrete emotions into six basic emotions (happiness, sadness, anger, fear, surprise, and disgust) \cite{ekman1987universals}, which are supported by cross-cultural researches showing that humans perceived these basic emotions in a similar form regardless of culture differences \cite{ekman1971constants}.   The dimensional emotion models propose that the emotional state can be accurately expressed as a combination of several psychological dimensions, such as the 2D ” circumplex model ” proposed by Russell \cite{russell1980circumplex} and the 3D dimensional model of Mehrabian et al. \cite{mehrabian1996pleasure}. In the widely adopted model proposed by Russell \cite{russell1980circumplex}, the valence dimension measures whether humans feel negative or positive, and the arousal dimension measures whether humans are bored or excited. Mehrabian et al. \cite{mehrabian1996pleasure} extended the emotional model from 2D to 3D by adding a dominance dimension, which measures submissive or empowered feelings.

\par The discrete emotion method is intuitive and widely used in peoples’ daily lives. However, it fails to cover the whole range of emotions exhibits by humans. The dimensional emotion method is less intuitive and often requires training the participants to use the dimensional emotion labelling system. Nevertheless, the dimensional emotion method is a more pragmatic and context-dependent approach to describe emotions \cite{zeng2008survey}. In this study, considering the primary emotions of drivers during driving, we combine both the discrete emotion and dimensional emotion methods to describe drivers’ negative emotions (e.g., anger) and positive emotions (e.g., happiness) quantitatively by employing the well-known emotion difference scale (DES) \cite{gross1995emotion} and self-assessed human body model (SAM) \cite{bradley1994measuring}.

\begin{table*}[]
\centering
\caption{The Summary of Reviewed publicly available datasets for facial  expression-based emotion recognition}
\begin{tabular}{|l|l|l|l|l|l|}
\hline
\textbf{dataset} & \textbf{\begin{tabular}[c]{@{}l@{}}\# of images/videos\\  and resolution\end{tabular}} & \textbf{Emotion} & \textbf{\# of participants} & \textbf{Condition} & \textbf{Emotion model/lables} \\ \hline
\begin{tabular}[c]{@{}l@{}}JAFFE\\ \cite{lyons1998japanese}\end{tabular} & \begin{tabular}[c]{@{}l@{}}213 images\\ $256\times256$\end{tabular} & \begin{tabular}[c]{@{}l@{}}Neutral, sadness, surprise,\\ happiness, fear, anger, \\ and disgust\end{tabular} & \begin{tabular}[c]{@{}l@{}}10 \\ 10 females\end{tabular} & \begin{tabular}[c]{@{}l@{}}Static life scenario\\ Controlled\\ Posed\end{tabular} & \begin{tabular}[c]{@{}l@{}}Discrete emotion model\\ - Neutral+6 basic emotions\end{tabular} \\ \hline
\begin{tabular}[c]{@{}l@{}}KDEF\\ \cite{lundqvist1998karolinska}\end{tabular} & \begin{tabular}[c]{@{}l@{}}4,900 images\\ $562\times762$\end{tabular} & \begin{tabular}[c]{@{}l@{}}Neutral, sadness, surprise,\\ happiness, fear, anger, \\ and disgust\end{tabular} & \begin{tabular}[c]{@{}l@{}}70\\ 35 females and 35 males\\ Age between 20 and 30\end{tabular} & \begin{tabular}[c]{@{}l@{}}Static life scenario\\ Controlled\\ Posed\end{tabular} & \begin{tabular}[c]{@{}l@{}}Discrete emotion model\\ -  Neutral+6 basic emotions\end{tabular} \\ \hline
\begin{tabular}[c]{@{}l@{}}MMI\\ \cite{pantic2005web}\end{tabular} & \begin{tabular}[c]{@{}l@{}}Over 2,900 \\ video sequences\\ $720\times576$\end{tabular} & \begin{tabular}[c]{@{}l@{}}Sadness, surprise, fear, \\ happiness, anger, \\ and disgust\end{tabular} & \begin{tabular}[c]{@{}l@{}}75\\ Age between 19 and 62\end{tabular} & \begin{tabular}[c]{@{}l@{}}Static life scenario\\ Controlled\\ Posed \& Spontaneous\end{tabular} & \begin{tabular}[c]{@{}l@{}}Discrete emotion model\\ -  6 basic emotions\end{tabular} \\ \hline
\begin{tabular}[c]{@{}l@{}}BU-3DFE\\ \cite{yin20063d}\end{tabular} & \begin{tabular}[c]{@{}l@{}}2,500 3D models\\ $1040\times1329$\end{tabular} & \begin{tabular}[c]{@{}l@{}}Neutral, sadness, surprise,\\ happiness, fear, anger, \\ and disgust\end{tabular} & \begin{tabular}[c]{@{}l@{}}100\\ 56 females and 44 males\\ Age between 18 and 70\end{tabular} & \begin{tabular}[c]{@{}l@{}}Static life scenario\\ Controlled\\ Posed\end{tabular} & \begin{tabular}[c]{@{}l@{}}Discrete emotion model\\ -  Neutral+6 basic emotions\\ -  4  levels of emotional intensity\end{tabular} \\ \hline
\begin{tabular}[c]{@{}l@{}}Multi-Pie\\ \cite{gross2010multi}\end{tabular} & \begin{tabular}[c]{@{}l@{}}755,370 images\\ $3072\times2048$\end{tabular} & \begin{tabular}[c]{@{}l@{}}Neutral, smile, surprise,\\ squint, disgust, \\ and scream\end{tabular} & \begin{tabular}[c]{@{}l@{}}337\\ 102 females and 235 males\end{tabular} & \begin{tabular}[c]{@{}l@{}}Static life scenario\\ Controlled\\ Posed\end{tabular} & \begin{tabular}[c]{@{}l@{}}Discrete emotion model\\ - Neutral+6emotion categories\end{tabular} \\ \hline
\begin{tabular}[c]{@{}l@{}}CK+\\ \cite{lucey2010extended}\end{tabular} & \begin{tabular}[c]{@{}l@{}}593 video \\sequences\\ $640\times480$,\\ $640\times490$\end{tabular} & \begin{tabular}[c]{@{}l@{}}Neutral, sadness, surprise,\\ happiness, fear, anger, \\ contempt, and disgust\end{tabular} & \begin{tabular}[c]{@{}l@{}}123\\ Age between 18 and 50\end{tabular} & \begin{tabular}[c]{@{}l@{}}Static life scenario\\ Controlled\\ Posed \& Spontaneous\end{tabular} & \begin{tabular}[c]{@{}l@{}}Discrete emotion model\\ - Neutral+7 emotion categories\end{tabular} \\ \hline
\begin{tabular}[c]{@{}l@{}}RaFD\\ \cite{langner2010presentation}\end{tabular} & \begin{tabular}[c]{@{}l@{}}8,040 images\\ $681\times1024$\end{tabular} & \begin{tabular}[c]{@{}l@{}}Neutral, sadness, surprise,\\ contempt, happiness, fear, \\ anger, and disgust\end{tabular} & \begin{tabular}[c]{@{}l@{}}67\\ 25 females and 42 males\end{tabular} & \begin{tabular}[c]{@{}l@{}}Static life scenario\\ Controlled\\ Posed\end{tabular} & \begin{tabular}[c]{@{}l@{}}Discrete emotion model\\ - Neutral+7 emotion categories\end{tabular} \\ \hline
\begin{tabular}[c]{@{}l@{}}DEAP\\ \cite{koelstra2011deap}\end{tabular} & \begin{tabular}[c]{@{}l@{}}880 video clips \\ (22 subjects)\\ $786\times576$\\Physiological \\signals\end{tabular} & \begin{tabular}[c]{@{}l@{}}Valence, arousal, \\ dominance\end{tabular} & \begin{tabular}[c]{@{}l@{}}32\\ 16 females and 16 males\\ Age between 19 and 33\end{tabular} & \begin{tabular}[c]{@{}l@{}}Static life scenario\\ Controlled\\ Spontaneous\end{tabular} & \begin{tabular}[c]{@{}l@{}}Dimensional emotion model\\ -  9  levels of valence, arousal, \\ dominance\end{tabular} \\ \hline
\begin{tabular}[c]{@{}l@{}}Belfast\\ \cite{sneddon2011belfast}\end{tabular} & \begin{tabular}[c]{@{}l@{}}1,400 video clips\\ $720\times576$ and\\ $1920\times1080$\end{tabular} & \begin{tabular}[c]{@{}l@{}}Disgust, fear, amusement,\\ frustration, surprise, \\ anger, and sadness\end{tabular} & \begin{tabular}[c]{@{}l@{}}256\\ 119 females and 137 males\end{tabular} & \begin{tabular}[c]{@{}l@{}}Static life scenario\\ Controlled\\ Natural tasks induced\end{tabular} & \begin{tabular}[c]{@{}l@{}}Discrete emotion model\\ - 7 emotion categories\\ -  emotional intensity\end{tabular} \\ \hline

\begin{tabular}[c]{@{}l@{}}DISFA\\ \cite{mavadati2013disfa}\end{tabular} & \begin{tabular}[c]{@{}l@{}}130,000 video \\ frames\\ $1024\times768$\end{tabular} & \begin{tabular}[c]{@{}l@{}}AU intensity for each \\ video frame (12 AUs)\end{tabular} & \begin{tabular}[c]{@{}l@{}}27\\ 12 females and 15 males\\ Age between 18 and 50\end{tabular} & \begin{tabular}[c]{@{}l@{}}Static life scenario\\ Controlled\\ Spontaneous\end{tabular} & 12AUs \\ \hline

\begin{tabular}[c]{@{}l@{}}RECOLA\\ \cite{ringeval2013introducing}\end{tabular} & \begin{tabular}[c]{@{}l@{}}3.8 hours videos\\ $1080\times720$\\Physiological,\\  audio signals\end{tabular} & Valence, arousal & \begin{tabular}[c]{@{}l@{}}46\\ 27 females and 19 males\\ Mean age 22\end{tabular} & \begin{tabular}[c]{@{}l@{}}Static life scenario\\ Controlled\\ Spontaneous\end{tabular} & \begin{tabular}[c]{@{}l@{}}Dimensional emotion model\\ - 9 levels of valence, arousal\end{tabular} \\ \hline

\begin{tabular}[c]{@{}l@{}}CFEE\\ \cite{du2014compound}\end{tabular} & \begin{tabular}[c]{@{}l@{}}5,060 images \\ $3000\times4000$\end{tabular} & \begin{tabular}[c]{@{}l@{}}22 categories of basic \\ and compound \\ emotions\end{tabular} & \begin{tabular}[c]{@{}l@{}}230\\ 130 females and 100 males\\ Mean age 23\end{tabular} & \begin{tabular}[c]{@{}l@{}}Static life scenario\\ Controlled\\ Posed\end{tabular} & \begin{tabular}[c]{@{}l@{}}Discrete emotion model\\ - 22 categories of basic \\ and compound emotions\end{tabular} \\ \hline
\begin{tabular}[c]{@{}l@{}}BP4D-\\ Spontaneous\\ \cite{zhang2014bp4d}\end{tabular} & \begin{tabular}[c]{@{}l@{}}328 sequences \\ of 3D+2D\\ $1040\times1329$\end{tabular} & \begin{tabular}[c]{@{}l@{}}Sadness, surprise, fear, \\ anger, embarrassment, \\ physical pain, happiness, \\ and disgust\end{tabular} & \begin{tabular}[c]{@{}l@{}}41\\ 23 females and 18 males\\ Age between 18 and 29\end{tabular} & \begin{tabular}[c]{@{}l@{}}Static life scenario\\ Controlled\\ Spontaneous\end{tabular} & \begin{tabular}[c]{@{}l@{}}Discrete emotion model\\ -  8 basic emotions\end{tabular} \\ \hline
\begin{tabular}[c]{@{}l@{}}ISED\\ \cite{happy2015indian}\end{tabular} & \begin{tabular}[c]{@{}l@{}}428 video \\sequences\\ $1920\times1080$\end{tabular} & \begin{tabular}[c]{@{}l@{}}Sadness, surprise, \\ happiness, and disgust\end{tabular} & \begin{tabular}[c]{@{}l@{}}50\\ 21 females and 29 males\\ Age between 18 and 22\end{tabular} & \begin{tabular}[c]{@{}l@{}}Static life scenario\\ Controlled\\ Spontaneous\end{tabular} & \begin{tabular}[c]{@{}l@{}}Discrete emotion model\\ -  8 basic emotions\\ -  5  levels of emotional intensity\end{tabular} \\ \hline
\begin{tabular}[c]{@{}l@{}}FER+\\ \cite{barsoum2016training}\end{tabular} & \begin{tabular}[c]{@{}l@{}}35,887 images\\ $48\times48$\end{tabular} & \begin{tabular}[c]{@{}l@{}}Neutral, surprise, sadness,\\ happiness, anger, disgust, \\ fear, contempt\end{tabular} & $\sim$35,887 & wild setting & \begin{tabular}[c]{@{}l@{}}Discrete emotion model\\ - Neutral+7 emotion categories\end{tabular} \\ \hline

\begin{tabular}[c]{@{}l@{}}EmotioNet\\ \cite{fabian2016emotionet}\end{tabular} & \begin{tabular}[c]{@{}l@{}}1,000,000 images\\ Various resolution\end{tabular} & \begin{tabular}[c]{@{}l@{}}23 basic or compound\\ emotions\end{tabular} & $\sim$100,000 & wild setting & \begin{tabular}[c]{@{}l@{}}Discrete emotion model\\ - 23 categories of basic and \\ compound emotions\end{tabular} \\ \hline

\begin{tabular}[c]{@{}l@{}}Aff-Wild\\ \cite{zafeiriou2017aff}\end{tabular} & \begin{tabular}[c]{@{}l@{}}298 video clips\\ Various resolution\end{tabular} & Valence and arousal & \begin{tabular}[c]{@{}l@{}}200\\ 70 females and 130 males\end{tabular} & wild setting & \begin{tabular}[c]{@{}l@{}}Dimensional emotion model\\ -  valecne and arousal\end{tabular} \\ \hline
\begin{tabular}[c]{@{}l@{}}RAVDESS\\ \cite{livingstone2018ryerson}\end{tabular} & \begin{tabular}[c]{@{}l@{}}7,356 video and \\ audio clips\\ $1280\times720$\end{tabular} & \begin{tabular}[c]{@{}l@{}}Neutral, calm, happiness,\\ sadness, anger, fear, \\ surprise, and disgust\end{tabular} & \begin{tabular}[c]{@{}l@{}}24\\ 12 females and 12 males\\ Age between 21 and 33\end{tabular} & \begin{tabular}[c]{@{}l@{}}Static life scenario\\ Controlled\\ Posed\end{tabular} & \begin{tabular}[c]{@{}l@{}}Discrete emotion model\\ -  Neutral+6 basic emotions\\ -  2  levels of emotional intensity\end{tabular} \\ \hline
\begin{tabular}[c]{@{}l@{}}AffectNet\\ \cite{mollahosseini2017affectnet}\end{tabular} & \begin{tabular}[c]{@{}l@{}}450,000 images \\ annotated manually\\ Various resolution\end{tabular} & \begin{tabular}[c]{@{}l@{}}Neutral, sadness, surprise,\\ happiness, fear, disgust, \\ anger, contempt\\ Valence, arousal\end{tabular} & 450,000 & wild setting & \begin{tabular}[c]{@{}l@{}}Discrete emotion model\\ - Neutral+7 emotion categories\\ Dimensional emotion model\\ -  valence and arousal\end{tabular} \\ \hline
\textbf{\begin{tabular}[c]{@{}l@{}}DEFE\\ (This \\ work)\end{tabular}} & \textbf{\begin{tabular}[c]{@{}l@{}}164 video clips\\ each 30s\\ $640\times480$,\\ $1920\times1080$\end{tabular}} & \textbf{\begin{tabular}[c]{@{}l@{}}Neutral, happiness, \\ anger\\ Valence, arousal, \\ dominance\end{tabular}} & \textbf{\begin{tabular}[c]{@{}l@{}}60\\ 13 females and 47 males\\ Age between 19 and 56\end{tabular}} & \textbf{\begin{tabular}[c]{@{}l@{}}Dynamic driving \\ scenarios\\ Controlled\\ Spontaneous\end{tabular}} & \textbf{\begin{tabular}[c]{@{}l@{}}Discrete emotion model\\ - 3 emotion categories\\ - 5  levels of emotional intensity\\ Dimensional emotion model\\ -  valence, arousal, dominance\\ -  9  levels of valence, arousal, \\ dominance\end{tabular}} \\ \hline
\end{tabular}
\end{table*}

\par Driver emotion recognition is often conducted by analyzing driver emotion expressions. The expressions of human emotions consists of facial expressions, speech, body posture and physiological changes. So far, different behavioural measurements (e.g., facial expression analysis, speech analysis, driving behaviour) \cite{wang2019feature}\cite{gao2014detecting}\cite{li2017estimation} , physiological signal measurements (e.g., skin electrical activity, respiration) \cite{wan2017road}\cite{lee2017wearable}, or self-reported scales (e.g., self-assessment manikin) \cite{malta2010analysis} have been applied in driver emotion recognition. Comparatively, physiological measurements are more objective and can be measured continuously. However, this measurement is highly invasive and may affect drivers’ driving performance. Self-reported measurements measure the subjective experience of drivers when applied correctly, but such measurements cannot take place during the study without interruption. For the study on the driver emotion in the driving environment, it is crucial to use non-invasive and non-contact measurement methods. High intrusiveness has a significant impact on both on the driver emotion expression and actual emotional experience, therefore should be avoided \cite{busso2004analysis}. To this end, this study employed facial expression to recognize driver emotions and ensure the continuity of data collection.

\par Facial expression is a powerful channel for drivers to express emotions \cite{yang2019facs3d}. Recent advances in facial expression-based emotion recognition have motivated the creation of multiple facial expression datasets. Publicly available datasets are fundamental for accelerating facial expression research. As shown in Table 1, we summarized the up-to-date representative public available datasets containing facial expressions. These datasets have been used for emotion recognition and to achieve different levels of success. As shown in Table I, one of the common aspects of these datasets is the collection of participants’ facial expression data in static life scenarios and wild settings. Although facial expression data collected in static life scenarios and wild settings can be employed to recognize emotions using various algorithms, it restricts the application of these algorithms into static life scenarios.

\par However, driving a car is a complex cognitive process \cite{groeger2000understanding}, which requires the driver to dynamically respond to driving tasks, such as visual cues, hazard assessment, decision-making, strategic planning\cite{li2019drivers}. Consequently, driving occupies a lot of driver’s cognitive resources \cite{lajunen2004manchester}, and cognitive processing is needed to elicit emotional responses \cite{brosch2013impact}. Driving affects drivers’ emotion expressions, which are different from the expressions in static life scenarios. As a result, if the above-mentioned algorithms are applied to dynamic driving scenarios, reliable recognition results may not be obtained. Thus, it is necessary to collect drivers’ facial expression data specifically for driver emotion recognition in dynamic driving scenarios and to analyze the human facial expression differences between dynamic driving scenarios and static life scenarios.

\section{DEFE Data Collection Framework}

\par To address the above-mentioned limitations, we introduce a driver emotion facial expression dataset (DEFE) in this study for driver emotion studies in intelligent vehicles. Table 2 presents the details of the experimental design for stimulus material selection, data collection, experiment protocol, and emotional labels. The performance of different emotion recognition algorithms was analyzed in this study. Also, this paper analyzed the differences in human facial expressions between dynamic driving scenarios and static life scenarios.

\begin{table}[]
\centering
\caption{Summary of DEFE dataset}
\begin{tabular}{ll}
\hline
\multicolumn{2}{l}{} \\
\multicolumn{2}{c}{\textbf{Video-audio stimulus selection}} \\
\multicolumn{2}{l}{} \\ \hline
Number of stimulus & 18 \\
Stimulus duration & 30s-120s \\
Initial stimulus selection & Manually selected \\
No. of rating per stimulus & 35 \\
Rating scales & \begin{tabular}[c]{@{}l@{}}Dimensional emotion model (SAM)\\ - Arousal\\ - Valence\\ - Dominance\\ Discrete emotion model (DES)\\ - Emotional categories\\ - Emotional intensity\end{tabular} \\
Rating values & Discrete scale of 1-9 \\
Selection method & \begin{tabular}[c]{@{}l@{}}Subset of annotated video-audio chips\\ with clearest and highest response\end{tabular} \\  \hline
\multicolumn{2}{l}{} \\
\multicolumn{2}{c}{\textbf{Driver facial expression data collection}} \\
\multicolumn{2}{l}{} \\ \hline
Number of participants & 60 (17females, 43 males) \\
Number of stimulus & 3 \\
Number of driving tasks & 3 \\
Rating scales & \begin{tabular}[c]{@{}l@{}}Dimensional emotion model (SAM)\\ - Arousal\\ - Valence\\ - Dominance\\ Discrete emotion model (DES)\\ - Emotional categories\\ - Emotional intensity\end{tabular} \\
Rating values & Discrete scale of 1-9 \\
Recorded signals & Driver facial expression videos \\ \hline
\multicolumn{2}{l}{} \\
\multicolumn{2}{c}{\textbf{DEFE dataset content}} \\
\multicolumn{2}{l}{} \\ \hline
Number of video clips & 164 \\
Emotions elicited & \begin{tabular}[c]{@{}l@{}}Anger (52 clips)\\ Happiness (56 clips)\\ Neutral (56 clips)\end{tabular} \\
Clip duration & 30s \\
Video clips format & MP4 \\
Image resolution & 1920*1080, 648*480 \\
Self-report of emotion & Yes \\
Emotion categories labels & \begin{tabular}[c]{@{}l@{}}3 categories\\ - Anger, happiness and neutral\end{tabular} \\
Emotion intensity labels & \begin{tabular}[c]{@{}l@{}}5  levels of anger and happiness\\ - 5 = no emotion, 9 = maximum intensity\end{tabular} \\
\begin{tabular}[c]{@{}l@{}}Valence, arousal, \\ dominance labels\end{tabular} & \begin{tabular}[c]{@{}l@{}}9  levels  of  valence, arousal, dominance \\ - 1  =  not at all, 9 = extremely\end{tabular} \\ \hline
\end{tabular}
\end{table}

\par In our DEFE dataset, video-audio clips were used as the stimuli to induce different emotions. To this end, a large number of video-audio clips were collected using a manual selection method. Subjective annotation was then performed to select the most appropriate stimulus material. Each stimulus material was labelled at least 35 times using SAM and DES scales, and the most effective three video-audio clips were selected to induce a specific driver emotion in our following experiments for data collection. Then, 60 drivers participated in the data collection experiment. After watching each of the three randomly sequenced video-audio clips selected to elicit a specific emotion, each participant completed the driving tasks in the same driving scenarios and rated their emotional responses in the driving process from the aspects of dimensional emotion and discrete emotion. Besides, we conducted classification experiments for the scales of arousal, valence, dominance, as well as the emotion category and intensity to establish the baseline results for our dataset in terms of classification accuracy and F1 score. Furthermore, we discussed the differences in facial expressions between dynamic driving scenarios and static life scenarios in the same culture by comparing the responses of different action units (AUs) in our DEFE and the JAFFE datasets.

\par The main contributions of this paper can be described as:
\begin{itemize}

\item We provide a new, publicly available dataset DEFE for spontaneous driver emotions analysis. The dataset contains frontal facial videos from 60 drivers, including their biographic information (gender, age, driving age), and subjective ratings on driver emotions (arousal, valence, dominance scales, as well as emotion category and intensity). To the best of our knowledge, this dataset is currently the only public dataset of driver facial expressions.

\item We compared the classification results of driver emotions on our DEFE dataset using the mainstream classification algorithms. The DEFE dataset supports driver emotion classification from two aspects, dimensional emotion (arousal, valence and dominance) and discrete emotion (emotional category and intensity). The comparisons established the baseline results of the introduced dataset with classification accuracy and F1 score.

\item The differences in human facial expressions between dynamic driving scenarios and static life scenarios were compared by analyzing drivers’ AUs presence, and the results showed significant differences between these two types of scenarios. Therefore, the previous human emotion datasets cannot be directly used for driver emotion analysis, and our introduced DEFE dataset fills this research gap.

\end{itemize}

\par The structure of this paper is as follows: Section III presents the selection of stimulus materials. Section IV introduces the DEFE data collection details, and the data processing, classification methods and results are described in Section V. Section VI compares human facial expression differences in dynamic driving scenarios and static life scenarios. The final conclusions are shown in Section VII.

\section{Video-audio Stimulus Selection}

\par The stimulus is necessary to elicit target emotions. All emotion datasets present the stimuli to evoke emotions, such as the international affective picture system (IAPS) \cite{lang1997international} and the international affective digitized sound system (IADS) \cite{bradley2007international}. Compared with images and music, videos and audios always bring strong emotional feelings. The existed researches have confirmed that video-audio clips can elicit the emotions of subjects reliably  \cite{gross1995emotion,schaefer2010assessing}  , hence video-audio clips were selected in our experiments. Eighteen initial video-audio clips were manually selected, and then we recruited participants to join a subjective rating experiment of these video-audio clips. Finally, three video-audio clips were selected based on the subjective rating results. Each of these steps is explained in detail as follows.

\subsection{Initial Video-Audio Clips Selection}

\par To select the most effective video-audio clips, two research assistants (1 male and 1 female) reviewed more than 500 video-audio clips and conducted the preliminary screening. They were asked to select video-audio clips that lasted 30-120 seconds and contained content to elicit a single target emotion, including a negative emotion (anger), a positive emotion (happy), and a neutral state. Another two research experts (1 male and 1 female) with rich experience in driver emotions analysis evaluated each selected video-audio clip. A consensus of the two experts decided the selections of the video-audio clips.

\par The selected video-audio clips are mainly based on Chinese real-life scenarios and events, such as aggressive driving and chatting. Other video-audio clips selection criteria include: 1) the video background should not be too dark, 2) the clip should contain complete speech segments, and 3) there is only one wanted expressing emotion in the clip. Accordingly, we selected 18 video-audio clips and checked them further in subjective annotation session.

\begin{table}[]
\centering
\caption{Brief Description of Selected Chinese Video-Audio stimulus}
\resizebox{0.48\textwidth}{!}{
\begin{tabular}{l|l|l}
\hline
\textbf{Targert Emotion} & \textbf{\begin{tabular}[c]{@{}l@{}}Duration (sec)\end{tabular}} & \textbf{Clip Content} \\ \hline
Happiness & 62 & \begin{tabular}[c]{@{}l@{}}Parents mentor their children to do\\  homework\end{tabular} \\ \hline
Anger & 45 & \begin{tabular}[c]{@{}l@{}}Many people were used in cruel \\ human experiments during the war\end{tabular} \\ \hline
Neutral & 48 & \begin{tabular}[c]{@{}l@{}}Man drives on city road with \\ nothing happened\end{tabular} \\ \hline
\end{tabular}}
\end{table}

\subsection{Subjective Annotation}

\par The web-based subjective emotion annotation experiment was conducted to evaluate the video-audio clips. For each participant, the 18 video clips were displayed in a random order, and there was a relatively long break time (3 minutes) between every two clips to avoid interference from the previous one. After watching each video-audio clip, participants finished two questionnaires based on their true feelings, namely the self- assessment manikin (SAM) \cite{bradley1994measuring} and the differential emotion scale (DES)\cite{gross1995emotion}.  SAM uses non-verbal graphical representations to assess the arousal, valence, and dominance level. The study in \cite{bradley1994measuring} has concluded the effectiveness of SAM. We adopted a 9-point scale (1 = ”not at all”, 9 = ” extremely ”) SAM \cite{bradley1994measuring} in our study for evaluation. DES is used to assess the different component of emotions, which consists of ten basic emotions. In this study, we used a 9-point scale DES (1= ”not at all”, 9= ” extremely ”) \cite{gross1995emotion} to assess the intensity of each self-reported emotional dimension. None of the clips was evaluated twice by the same participant, and at least 35 assessments were collected for each video.

\begin{figure*}[!t]

\centering
\includegraphics[width=0.98\textwidth]{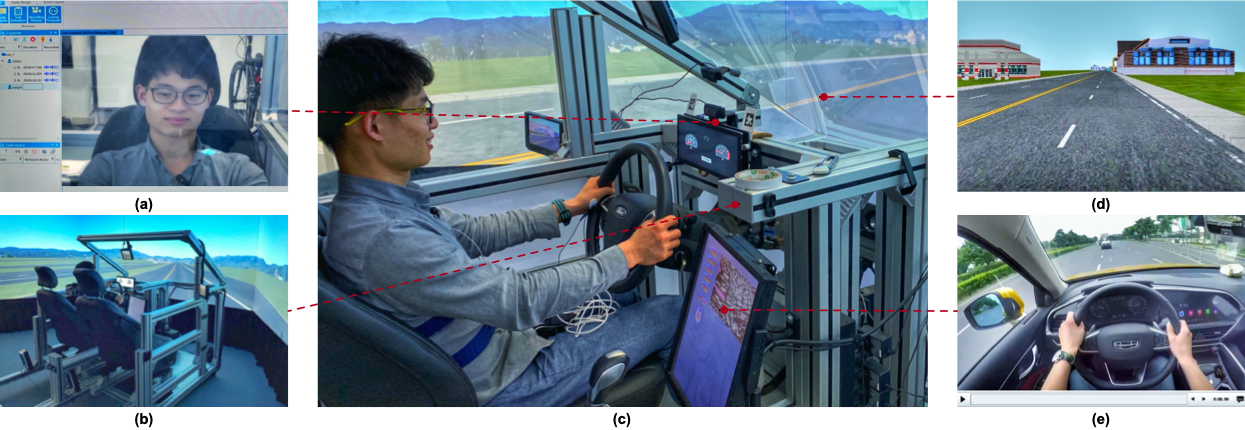}
\caption{Experimental setup of driver facial expression data collection. (a) driver facial expression recording, (b) fix-based driving simulator, (c) experiment setup, (d) driving scenarios, (e) video-audio stimulus display}
\label{fig:label1}
\end{figure*}

\subsection{Video-Audio clips Selection}

\par Three video-audio clips were selected by comprehensively considering the SAM and DES results. Firstly, we normalized the variables by calculating the Z-scores and then conducted a cluster analysis using the K-means algorithm to identify the clusters of emotions based on the SAM data. The clustering results showed that a total of three emotion categories were generated, which corresponded to the positive emotion (happy), negative emotion (anger), and neutral, respectively. The video-audio clip whose rating was closest to the extreme corner of each quadrant was selected and marked as the representative video-audio clip of the cluster \cite{koelstra2011deap}.

\par Moreover, we selected video-audio clips for each emotional category based on the following scores of the DES data: 1) Hit rate was defined as the proportion of participants who chose the target emotion. 2) The intensity value was defined as the average score of target emotions. 3) The success index represented the sum of the two Z-scores, which were obtained by normalizing the hit rate and intensity values. Next, video-audio clips with the highest success index were selected from each emotion category, and representative video-audio clips were also selected according to the SAM data for verification. It should be noted that the neutral video-audio clip was only selected based on the clustering results. Eventually, as shown in Table 3, three of the most effective videos were selected for driver facial expression data collection experiment.

\section{Driver Facial Expression Data Collection}

\subsection{Ethics Statement}
\par The experimental procedure and the video content shown to the participants were approved by Chongqing University Cancer Hospital Ethics Committee, China. Participants and data from participants were treated according to the Declaration of Helsinki. The participants were also informed that they had the right to quit the experiment at any time. The video recordings of the participants were included in the dataset only after they gave written consent for the use of their videos for research purpose. A few participants were also agreed to use their face images in research articles.

\subsection{Participants}

\par Sixty Chinese participants(47 males and 13 females) with aging from 19 to 56 years old (mean [M] = 27.3 years, standard deviation [SD] = 7.7. Years) were recruited to participate in this experiment from Shapingba District, Chongqing, China. Each participant had a valid driving license with at least one year of driving experience (average [M] = 5.5 years, standard deviation [SD] = 5.8, range = 1-30 years). All the participants had normal or corrected to normal vision (36 participants wear glasses) and normal hearing ability. The presence of occlusion such as glasses is a significant research challenge of facial expression recognition; hence participants wearing glasses were included to evaluate the robustness of emotion recognition. All the participants signed the consent form to participate in the study and received 100 CNY as financial reimbursement for their participation.

\subsection{Experiment Setup}

\par The experiments were carried out in a fix-based driving simulator (Figure.\ref{fig:label1}(b)) with illumination-controlled (RDS2000, Real-time technology SimCreator, Ann Arbor, Michigan, USA). Figure.\ref{fig:label1}(d) shows the front view, which was presented using three projectors, and the rear view was displayed using three LCD screens (one for the rearview in the vehicle and two for the left and right rear views). Another two LCD screens were used to display the dashboard and central stackf. The ambient noise and sound of the engine were presented through two speakers. The vibration of the vehicle was simulated through a woofer under the driver’s seat. For the presentation of stimulus without changing the internal environment of the driving simulator, as shown in Figure.\ref{fig:label1}(e), we used a 20-inch central stack screen ($1,280\times1,024$, 60Hz) to display the video-audio stimulus materials. A stereo Bluetooth speaker (Xiaomi) was used to play the audio, and the audio volume was set to a relatively loud level. However, each participant was asked before the experiment whether the volume was comfortable and adjusted when necessary for clear hearing. During the experiment, as shown in Figure.\ref{fig:label1}(a),   the participants’ faces were continuously imaged with a visual camera. The visual face camera was an HD Pro Webcam C920 (Logitech, Newark, CA.) with a resolution of $1,920\times1,080$ pixels, collecting data at a frame rate of 30 fps. Also, an iPad (Apple) was used for participant self-reported emotion. Figure.\ref{fig:label1}(c) shows the overall data collection experiment setup.

\begin{figure}[!t]
\centering
\includegraphics[width=0.48\textwidth]{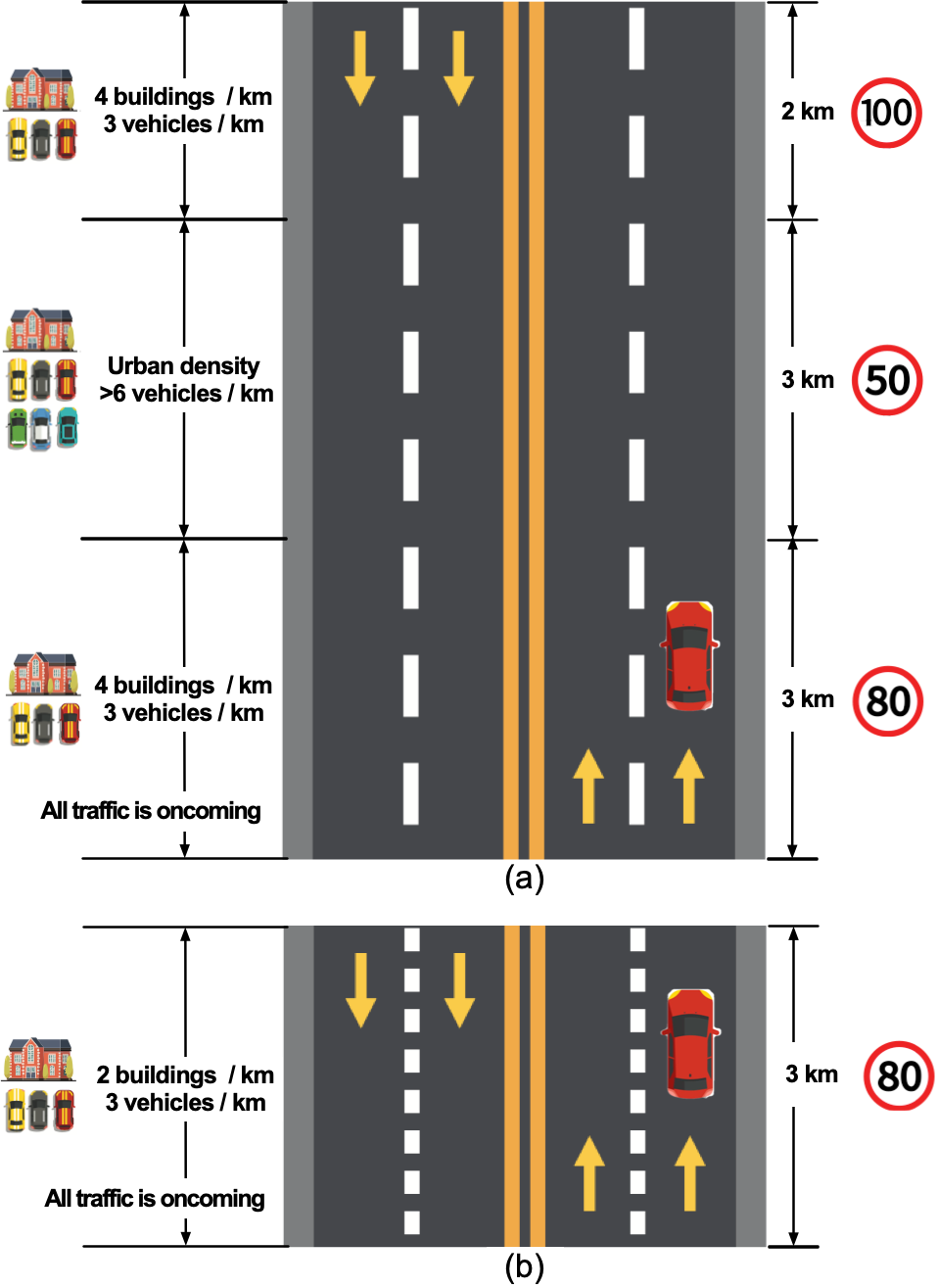}
\caption{Driving scenarios and tasks. (a) practice driving, (b) emotional driving }
\label{fig:label2}
\end{figure}

\subsection{Driving Scenarios and Tasks}

\par Two driving scenarios on highways were realized in the simulator. The reason for setting these two scenarios is to minimize the impact of complex driving scenarios on driver performance. The first was a practice scenario (PD) to help participants familiarize themselves with the simulator before the experiment. As shown in Figure.\ref{fig:label2}(a), the practice scenario was an 8km straight section of a four-lane highway with two for each driving direction. The participants were asked to drive on the right lane with the speed changes in the range of 80km/h – 50km/h – 100km/h. The second scenario is an emotional driving (ED) scenario. As shown in Figure.\ref{fig:label2}(b), the emotional driving scenario was a 3km straight section of the same highway with a posted speed limit of 80km/h. The participants were asked to drive on the right lane with speed around 80km/h.

\subsection{Experiment protocol}

\par
\begin{figure}[!t]
\centering
\includegraphics[width=0.48\textwidth]{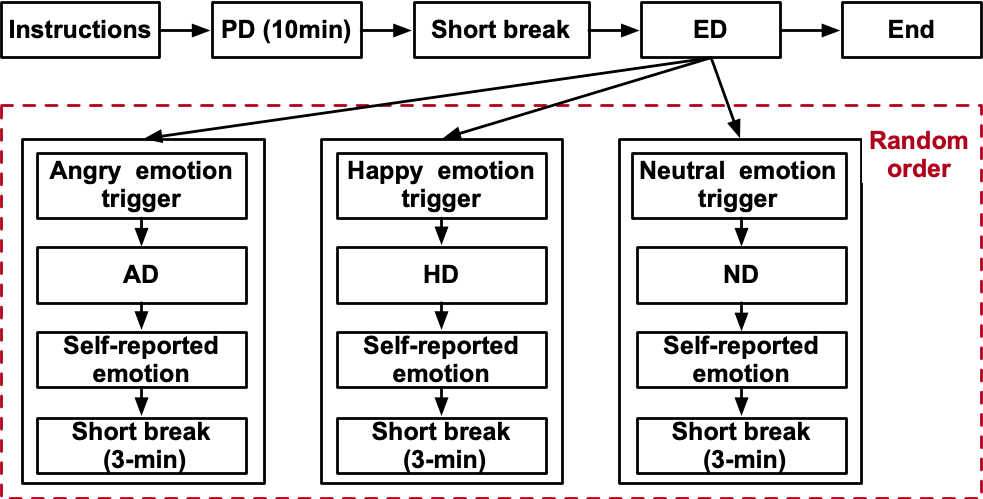}
\caption{Experiment protocol.}
\label{fig:label3}
\end{figure}

 To obtain drivers’ ED data, we designed an experimental protocol about 45 minutes driving. The protocol was composed of one PD, followed by three ED. ED driving included angry driving (AD), happy driving (HD) and neutral driving (ND). Figure.\ref{fig:label3} presented details of the protocol. Before the experiment, each participant signed a consent form and filled out a basic information questionnaire (gender, age, driving age). Next, they were provided with a set of instructions to inform them of the experimental protocol and the definition of different scales used for self-reported emotions. Then, the participants were required to drive a 10-minute PD to help them get familiar with the operation and motion performance of the driving simulator. After a short break following PD, the participants started the three EDs. The corresponding emotion was induced by watching the selected video-audio clip at the beginning of each ED, following by driving with emotion. At the end of each ED session, the participant was required to report his/her self-evaluated emotion level using SAM and DES. There was a 3 minutes break between each two EDs. During the entire experiment, if the participants felt any discomfort, they could withdraw from the experiment at any time.

\subsection{Self-Reported Emotion}

\par To identify the emotion experienced by participants,  we employed self-reported scales for subjective assessment of emotions. After each driving task, the participants were asked to assess their emotional experience while driving using SAM and DES. The SAM and DES scales were presented to participants by an iPad. In SAM, the valence scale ranged from unhappy to happy, the arousal scale ranged from calm to stimulation, and the dominance scale ranged from submissive (or ”without control”) to dominant (or ”under control, empowered ”). In DES, there were ten emotion dimensions, and each dimension evaluated the intensity of emotions from ”not at all” to ” extremely ”. Each dimension of the SAM scale and the DES scale is represented from one to nine by a Likert scale. If the self-assessments from participants were not consistent with the induced target emotions, we would use the participants’ self-reported data as the ground truth to label the facial video data.

\section{Data Processing and Evaluation}

\subsection{Data Processing}

\par In this section, we described the processing of driver facial expression data. First, we labelled the facial expression data of 60 drivers according to their self-reported emotion and removed the ED data that was not successfully induced. Second, we reported how to split data for driver emotion recognition, including splitting effective video clips from the original data and extracting driver facial expression.

\begin{figure}

\centering
\includegraphics[width=0.48\textwidth]{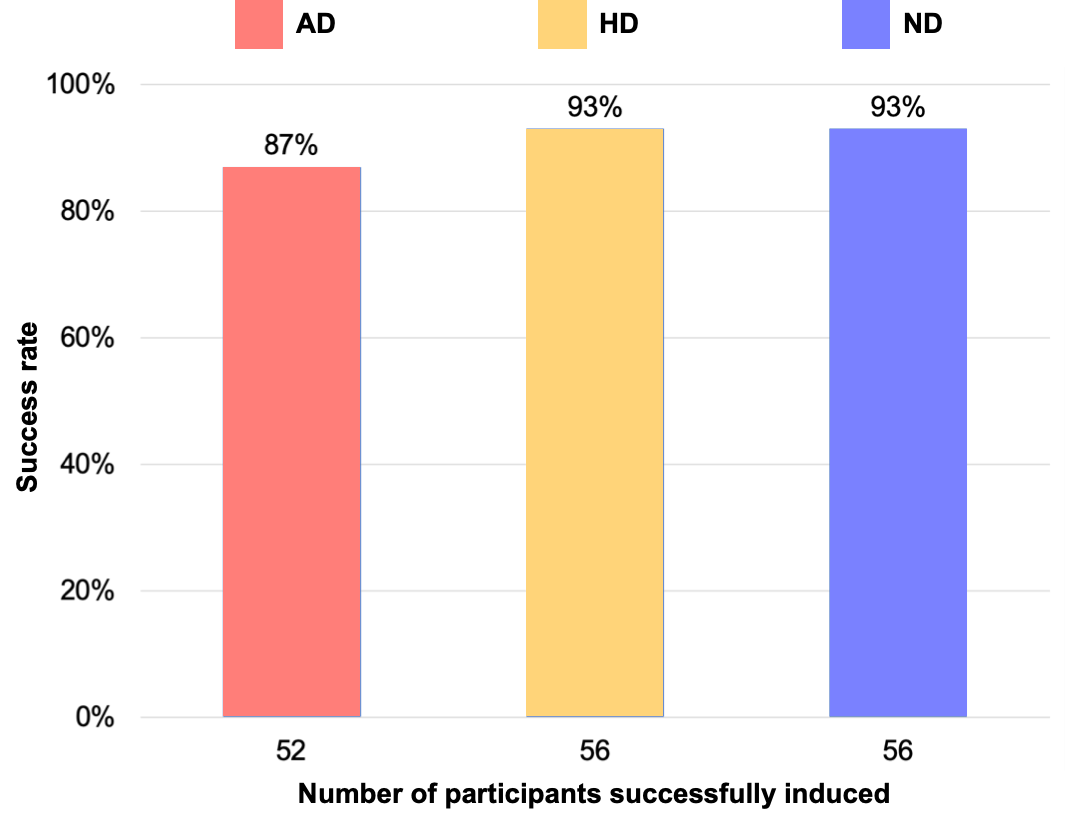}
\caption{The emotional driving induction success rate for 60 drivers.}
\label{fig:label4}
\end{figure}

\par During data collection, each participant completed three ED sessions with average recording data of 405s. Also, we compiled the self-reported data for each participant. As shown in Figure.\ref{fig:label4}, the numbers of successfully induced emotional drivers were 52, 56, and 56 for the anger, happy, and neutral driving, respectively. Participants’ self-reported data were used as the ground truth to label driver facial expression data.

\begin{figure*}[!t]

\centering
\includegraphics[width=0.98\textwidth]{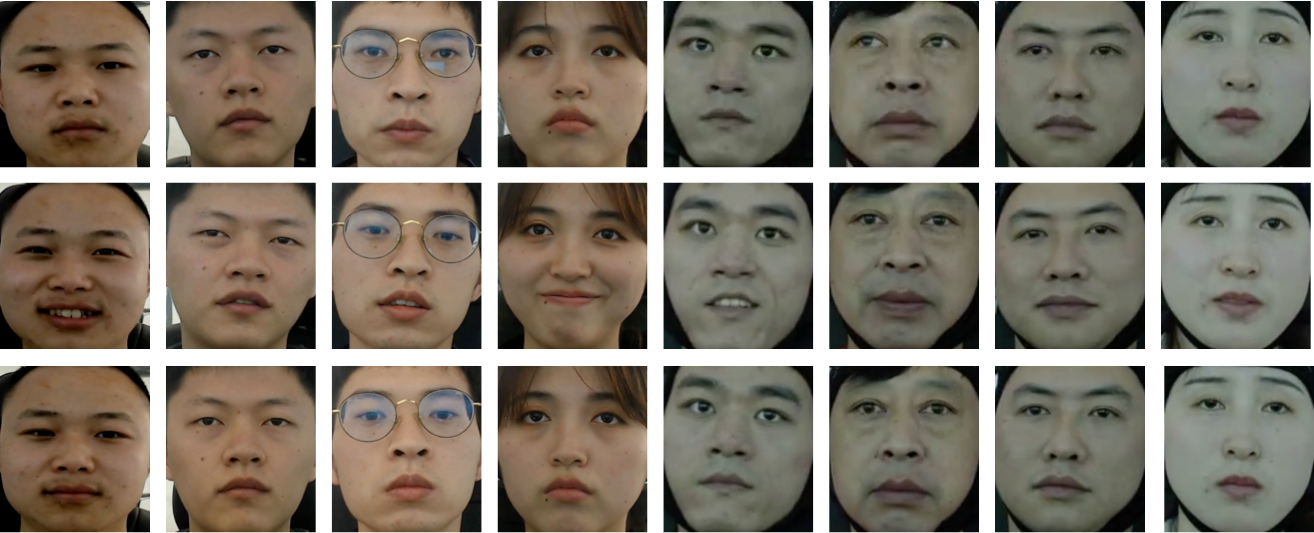}
\caption{Sample images of the 3 emotion categories in DEFE dataset, anger (1st row), happiness (2nd row) and neutral (3rd row).}
\label{fig:label5}
\end{figure*}

\par As per \cite{bos2006eeg} and \cite{levenson1991emotion}, the facial expression video sequences 15s after drivers started driving were clipped as the most effective data. Face detection and alignment in driving environments are challenging due to various poses, illuminations and occlusions (glasses). MTCNN (Multi-task Cascaded Convolutional Networks) is a cascade structure based on deep learning, which is relatively accurate when detecting faces in multiple pose angles and in unconstrained scenes \cite{zhang2016joint}. Hence, we used MTCNN to track and extract driver face data from each video frame. After extracting driver face expression data, we obtained a total of 17,310 image frames of driver faces with 64*64 pixel.

\par Therefore, the created dataset contains facial expression videos and images from 60 drivers with the ground truth of dimensional emotion (valence, arousal and dominance) and discrete emotion (emotion categories and its intensity). A few examples of the dataset images are provided in Figure.\ref{fig:label5}, which shows that drivers’ facial expressions varied with the types of emotion, but the variation was weak in some cases during driving, for example, the difference between AD and ND was tiny. Most video clips were challenging to observe peak expressions, and we also observed that the change of emotion with driving duration was weak, and this phenomenon is probably because the facial expression of emotion was affected by driving tasks.

\subsection{Classification Protocol}

\par In this section, we introduced two different types of protocols for driver emotion recognition based on facial expression data. (1) To investigate driver emotion classification results based on the dimensional emotion model, we proposed three different nine-classification problems: valence, arousal, and dominance. To this end, the SAM scores of participants were used as the ground truth. Each classification (valence, arousal, dominance) on these scales was divided into nine levels (1 = ”not at all”, 9 = ”extremely”). (2) To study driver emotion classification results based on the discrete emotion model, we proposed a three-emotion classification protocol, namely anger, happiness, and neutral. Besides, we discussed the intensity recognition for anger and happy emotions, respectively. To this end, the DES scores were taken as the ground truth. Each emotion(anger and happiness) intensity was divided into 5 levels (5 = ”no emotion”, 9 = ”maximum intensity”).

\par It should be noted that the above approach can lead to unbalanced classes for some participants and scales. In light of this, we included F1 scores in order to report reliable results. The F1 score is a commonly used metric in classification tasks, which considers both precision (P) and recall (R) of the model. It quantifies the correct prediction of the positive samples. When categories are unbalanced, the F1 score will be attenuated \cite{jeni2013facing}. We additionally used accuracy as another metric. Accuracy quantifies how well the classification correctly identifies or excludes conditions, and it is robust to unbalanced data.

\par Both the traditional and the deep learning methods for emotion recognition tasks were included in this study. As the most effective traditional method in most classification tasks \cite{lucey2010extended}, SVM (Support Vector Machine) was selected to be implemented by the sklearn toolbox with a linear kernel. As for the deep learning-based classification methods, Xception \cite{arriaga2017real} was applied. The Xception network has been widely adopted in emotion recognition tasks, and many state-of-the-art emotion recognition networks are developed based on the Xception network\cite{pramerdorfer2016facial}\cite{li2018deep}. For the network, the loss function can be expressed as:

\begin{equation}\label{eq1}
L(y, \hat{y})=-\sum_{j=0}^{M} \sum_{i=0}^{N}\left(y_{i j} * \log \left(\hat{y}_{i j}\right)\right)
\end{equation}

Equation\ref{eq1} where $\hat{y}$ is the prediction and $y$ is the ground truth. The above deep learning method used the same training strategy. First, it employed Adam optimizer \cite{kingma2014adam}, which has a learning rate of $10^{-3}$ and a weight decay of $10^{-6}$ for training. Second, image augmentations, including random horizontal flips, random crop, and random rotation, were applied on-the-fly to increase the amount of training images effectively. SVM was applied with Intel R CoreTM i5-dual-core CPU. Xception was used with TITAN XP.

\subsection{Evaluation Results}

\par Apart from the emotion recognition results for the proposed dataset, we also selected the DEAP \cite{koelstra2011deap} and CK + \cite{lucey2010extended} datasets which were collected in static life scenarios as the comparison datasets. The DEAP dataset consists of 32 participants. Each participant watched 40 1-minute long video-audio chips as the emotional stimulus while recording facial videos and physiological signals. There are 40 trials recorded per participant, each corresponding to one emotion elicited by one video-audio chip. After watching each video, the participants were asked to assess their real emotions from five dimensions: valence, arousal, dominance, liking and familiarity. The rating ranges from 1 (weakest) to 9 (strongest), except liking and familiarity, which rating from 1 to 5. Facial videos from 22 of the participants were also recorded at the same time. This paper adopted the 22 facial videos in this dataset and investigated the emotion classification results based on the dimensional emotion model for comparison. The CK + dataset consists of 123 participants. This dataset was posed and spontaneous by multiple participants whose facial expressions started from neutral to the peak. In the CK + dataset, 327 sequences have discrete emotion labels including neutral, sadness, surprise, happiness, fear, anger, contempt and disgust. This paper selected the neutral, anger and happy sequences in this dataset to compare the emotion classification results based on discrete emotion models.

\begin{table}[]
\centering
\caption{Average Accuracies (ACC) and F1-Scores (F1, Average Score for each class) in protocol one based on the dimensional emotion model(in \%).}
\resizebox{0.47\textwidth}{!}{
\begin{tabular}{|l|l|l|l|l|l|l|l|}
\hline
\multicolumn{1}{|c|}{\multirow{2}{*}{\textbf{dataset}}} & \multicolumn{1}{c|}{\multirow{2}{*}{\textbf{Method}}} & \multicolumn{2}{c|}{\textbf{Valence}} & \multicolumn{2}{c|}{\textbf{Arousal}} & \multicolumn{2}{c|}{\textbf{Dominance}} \\ \cline{3-8} 
\multicolumn{1}{|c|}{} & \multicolumn{1}{c|}{} & \multicolumn{1}{c|}{\textbf{ACC}} & \multicolumn{1}{c|}{\textbf{F1}} & \multicolumn{1}{c|}{\textbf{ACC}} & \multicolumn{1}{c|}{\textbf{F1}} & \multicolumn{1}{c|}{\textbf{ACC}} & \multicolumn{1}{c|}{\textbf{F1}} \\ \hline
\multirow{2}{*}{\textbf{DEFE}} & SVM & 53.39 & 54.79 & 59.49 & 63.04 & 59.49 & 63.04 \\ \cline{2-8} 
 & Xception & 86.00 & 83.73 & 91.54 & 91.76 & 88.17 & 79.55 \\ \hline
\multirow{2}{*}{\textbf{DEAP}} & SVM & 27.88 & 23.24 & 29.82 & 23.25 & 28.12 & 24.14 \\ \cline{2-8} 
 & Xception & 24.10 & 21.41 & 35.06 & 31.80 & 31.00 & 24.24 \\ \hline
\end{tabular}}
\end{table}

Table 4 shows the average accuracies and F1 scores (average F1 scores for nine classes) for each rating scale (valence, arousal and dominance) when using protocol one on DEFE. We compared the performances of SVM and Xception on the DEFE dataset. In general, the accuracies when using Xception method were at least $30\%$ higher than the accuracy when using SVM. The highest classification accuracy for valence, arousal, and dominance achieved $86.00\%$, $91.54\%$, and $88.17\%$, respectively, when using Xception. In terms of the F1 scores, the highest scores for valence, arousal, and dominance were: $83.73\%$, $91.76\%$, and $79.55\%$ respectively, when using Xception. In addition to the emotion recognition results on the DEFE dataset, Table VI also shows the comparison results on the DEAP dataset when using the same recognition algorithms. The results show that the DEFE dataset had higher recognition accuracies and F1 scores than the DEAP dataset, which may be because the participants’ faces were affixed with electrode pads for physiological signals collection in the DEAP dataset, which affected the facial expression recognition results.

\begin{table}[]
\centering
\caption{Average Accuracies (ACC) and F1-Scores (F1, Average Score for each class) in protocol two based on the discrete emotion model(in \%).}
\resizebox{0.5\textwidth}{!}{
\begin{tabular}{|l|l|l|l|l|l|l|l|}
\hline
\multirow{2}{*}{\textbf{Dataset}} & \multirow{2}{*}{\textbf{Method}} & \multicolumn{2}{l|}{\textbf{Emotion category}} & \multicolumn{2}{l|}{\textbf{Angry intensity}} & \multicolumn{2}{l|}{\textbf{Happy intensity}} \\ \cline{3-8} 
 &  & \textbf{ACC} & \textbf{F1} & \textbf{ACC} & \textbf{F1} & \textbf{ACC} & \textbf{F1} \\ \hline
\multirow{2}{*}{\textbf{DEFE}} & SVM & 53.08 & 52.93 & 86.01 & 87.42 & 85.41 & 85.57 \\ \cline{2-8} 
 & Xception & 90.34 & 90.21 & 97.60 & 97.12 & 97.88 & 97.59 \\ \hline
\multirow{2}{*}{\textbf{CK+}} & SVM & 82.70 & 71.45 & \multicolumn{1}{c|}{-} & \multicolumn{1}{c|}{-} & \multicolumn{1}{c|}{-} & \multicolumn{1}{c|}{-} \\ \cline{2-8} 
 & Xception & 94.31 & 93.25 & \multicolumn{1}{c|}{-} & \multicolumn{1}{c|}{-} & \multicolumn{1}{c|}{-} & \multicolumn{1}{c|}{-} \\ \hline
\end{tabular}}
\end{table}

\par Similarly, Table 5 shows the average accuracies and F1 scores for the emotion categories (anger, happiness, and neutral) when using protocol two. We also compared the classification results when using SVM, and Xception in Table V. The results show that both the highest classification accuracy ($90.34\%$) and the highest F1 scores ($90.21\%$) were obtained when using Xception. Apart from the emotion recognition results on DEFE, Table V also presented the comparison results on the CK + dataset when using the same recognition algorithms. The results show that the recognition results of the CK + dataset were higher than that of DEFE dataset.

\par Moreover, Table 5 shows the average accuracies and F1 scores of the intensity classification results on anger and happiness emotions when using protocol two with different algorithms. Five classes of the intensity of anger and happiness were classified based on facial expression data. The results show that the highest classification accuracies for angry and happy driving intensity were $97.60\%$ and $97.88\%$, respectively. The highest F1 scores for angry and happy intensity were $97.88\%$ and $97.59\%$, respectively. It should be noted that in recognition of emotion intensity, we did not compare the results with other datasets, because there was currently no spontaneous facial expression datasets with emotional intensity labels.

\par The comparison results in this section show that there is a difference in human facial expression between DEFE and CK+. Due to the influence of driving tasks in driving scenarios, facial expressions of drivers may be suppressed when they experience emotional states. Hence, it is necessary to discuss further the difference between human facial expressions in dynamic driving scenarios and static life scenarios.

\section{The facial expression difference between dynamic driving and static life conditions}

\subsection{dataset Selection for Comparison}

\par In this section, we conducted a differential analysis of the facial expressions between dynamic driving and static life conditions by comparing the DEFE and JAFFE datasets. The static life dataset, Japanese Female Facial Expression (JAFFE) dataset\cite{lyons1998japanese}, was selected as a baseline. It was posed by 10 East-Asian females with seven emotion expressions (happy, anger, disgust, fear, sad, and neutral). Each female had two to four examples for each emotion. In total, there are 213 grayscale facial expression images in this dataset.

\par Given the East-Asian cultural background with small difference, the JAFFE dataset was the most optimal control group for our DEFE dataset because of the excluded most cultural bias\cite{Jack7241}. Since DEFE only include two emotions (anger and happiness), we also selected anger and happiness expressions from JAFFE for analysis.  Meanwhile, gender differences may affect the results so that we removed the male drivers from the initial DEFE dataset. 

\begin{figure}[!t]
\centering
\includegraphics[width=0.3\textwidth]{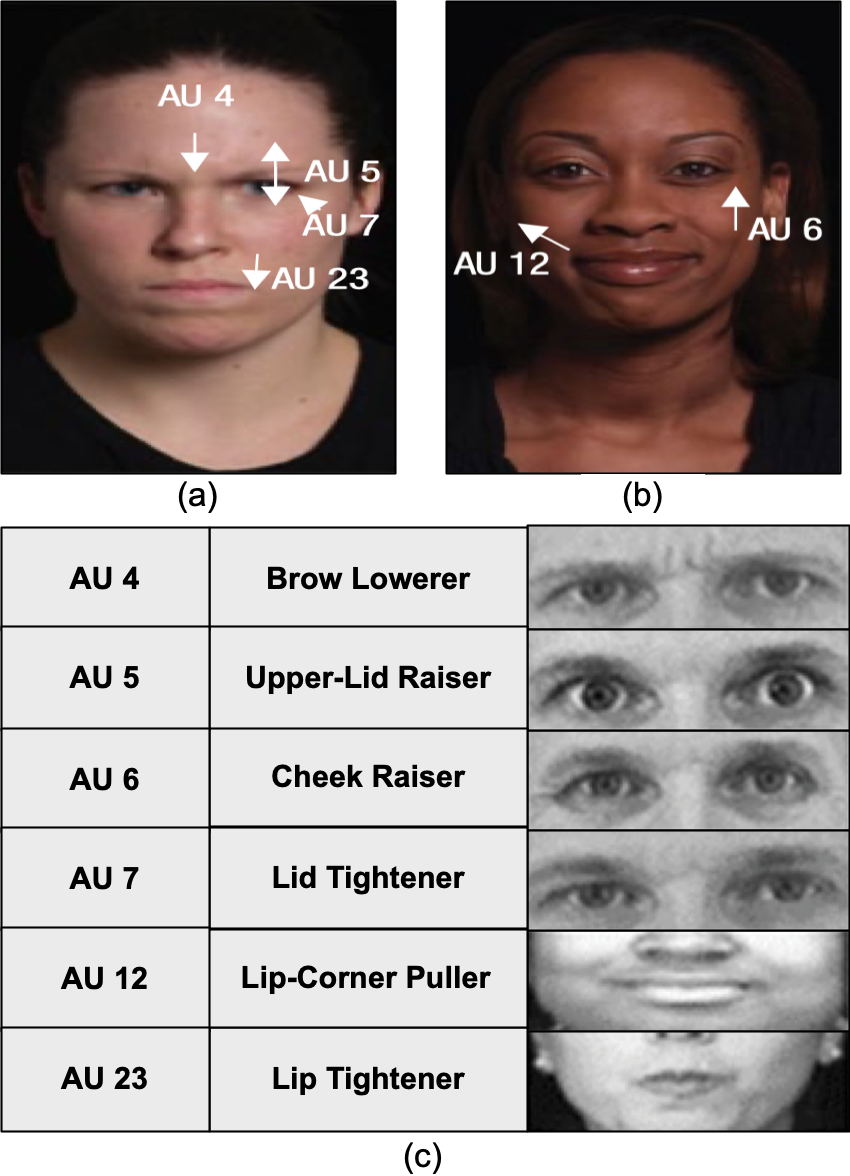}
\caption{Facial action coding system (FACS) codes can be used to describe the facial configuration in adults. (a) and (b) display the common FACS codes for anger and happiness, respectively, and (c) presents the AUs description for anger and happiness \cite{barrett2019emotional}}
\label{fig:label6}
\end{figure}

\subsection{Differential Analysis Protocol}

Each participant’s facial expressions were evaluated by observing subtle changes in facial features. The Facial Action Coding System (FACS) \cite{ekman_facial_2002} is a systematic approach to describe what a face looks like when facial muscle movements have occurred. There are 44 coded facial muscle movements, namely Action Units (AUs), in FACS according to the presence and intensity of facial movements. Ekman et al. further proposed that facial emotion expressions could be coded as a combination of several AUs. Figure.\ref{fig:label6} (a) and (b) display the common FACS \cite{barrett2019emotional} codes for anger and happiness, respectively, and Figure.\ref{fig:label6} (c) presents the AUs descriptions for anger and happiness. In this study, the AU codes for anger (AU 4, 5, 7 and 23) and happiness (AU 6 and 12) were used as the basic units for differential analysis.

\par We utilized OpenFace \cite{baltruvsaitis2016openface}, a facial expression analysis toolkit, to detect the presence of AUs. When an AU was detected, we coded it as 1 and otherwise 0. Due to video enable to capture enriched data, DFEE contained more facial expression information compared than JAFFE. In the end, the number of observations of happy and anger expressions in JAFFE was 61. DEFE, as a video dataset, had 10020 and 6660 number of observations of happy and angry expressions.

\par To analyze the differences of AU presence between dynamic driving and static life conditions, we conducted a statistical analysis to investigate the presence of AUs in the two datasets. Given the same emotions in both datasets, we should not observe a statistical difference if the facial expressions were similar between dynamic driving and static life conditions. Meanwhile, the average difference between the two datasets may not fully reflect emotional changes. Instead, it may be led by the baseline difference of two datasets.

\par Hence, to study the relationship of these AUs to anger and happiness in the two datasets,  a logit regression was performed on the two datasets separately with happiness coded as 1 and anger as 0. If the relationship coefficients of AUs had differences in the two datasets, it could be concluded that some AUs performed differently between dynamic driving and static life scenarios. It should be noted that positive coefficient means the AU is related to happiness and negative coefficient means the AU is related to anger.

\begin{table*}[]
\centering
\caption{Statistics analysis results of AUs' presence in Anger and Happiness cross DEFE and JAFFE dataset}
\begin{tabular}{lllllll}
\hline
\multicolumn{3}{|l|}{\textbf{The presence of AUs in anger}} & \multicolumn{1}{l|}{\textbf{AU 4}} & \multicolumn{1}{l|}{\textbf{AU 5}} & \multicolumn{1}{l|}{\textbf{AU 7}} & \multicolumn{1}{l|}{\textbf{AU 23}} \\ \hline
\multicolumn{1}{|l|}{\multirow{5}{*}{\textbf{Anger}}} & \multicolumn{1}{l|}{\multirow{2}{*}{JAFFE}} & \multicolumn{1}{l|}{Average} & \multicolumn{1}{l|}{0.433} & \multicolumn{1}{l|}{0.683} & \multicolumn{1}{l|}{0} & \multicolumn{1}{l|}{0.05} \\ \cline{3-7} 
\multicolumn{1}{|l|}{} & \multicolumn{1}{l|}{} & \multicolumn{1}{l|}{STD} & \multicolumn{1}{l|}{0.5} & \multicolumn{1}{l|}{0.469} & \multicolumn{1}{l|}{0} & \multicolumn{1}{l|}{0.22} \\ \cline{2-7} 
\multicolumn{1}{|l|}{} & \multicolumn{1}{l|}{\multirow{2}{*}{DEFE}} & \multicolumn{1}{l|}{Average} & \multicolumn{1}{l|}{0.066} & \multicolumn{1}{l|}{0.351} & \multicolumn{1}{l|}{0.467} & \multicolumn{1}{l|}{0.157} \\ \cline{3-7} 
\multicolumn{1}{|l|}{} & \multicolumn{1}{l|}{} & \multicolumn{1}{l|}{STD} & \multicolumn{1}{l|}{0.248} & \multicolumn{1}{l|}{0.477} & \multicolumn{1}{l|}{0.499} & \multicolumn{1}{l|}{0.364} \\ \cline{2-7} 
\multicolumn{1}{|l|}{} & \multicolumn{2}{c|}{T-test} & \multicolumn{1}{l|}{5.689***} & \multicolumn{1}{l|}{5.464***} & \multicolumn{1}{l|}{-76.378***} & \multicolumn{1}{l|}{-3.733***} \\ \hline
\multicolumn{3}{|l|}{\textbf{The presence of AUs in Happiness}} & \multicolumn{1}{l|}{\textbf{AU 6}} & \multicolumn{1}{l|}{\textbf{AU 12}} & \multicolumn{1}{c|}{-} & \multicolumn{1}{c|}{-} \\ \hline
\multicolumn{1}{|l|}{\multirow{5}{*}{\textbf{Happiness}}} & \multicolumn{1}{l|}{\multirow{2}{*}{JAFFE}} & \multicolumn{1}{l|}{Average} & \multicolumn{1}{l|}{0.361} & \multicolumn{1}{l|}{0.475} & \multicolumn{1}{c|}{-} & \multicolumn{1}{c|}{-} \\ \cline{3-7} 
\multicolumn{1}{|l|}{} & \multicolumn{1}{l|}{} & \multicolumn{1}{l|}{STD} & \multicolumn{1}{l|}{0.484} & \multicolumn{1}{l|}{0.504} & \multicolumn{1}{c|}{-} & \multicolumn{1}{c|}{-} \\ \cline{2-7} 
\multicolumn{1}{|l|}{} & \multicolumn{1}{l|}{DEFE} & \multicolumn{1}{l|}{Average} & \multicolumn{1}{l|}{0.177} & \multicolumn{1}{l|}{0.18} & \multicolumn{1}{c|}{-} & \multicolumn{1}{c|}{-} \\ \cline{2-7} 
\multicolumn{1}{|l|}{} & \multicolumn{1}{l|}{} & \multicolumn{1}{l|}{STD} & \multicolumn{1}{l|}{0.382} & \multicolumn{1}{l|}{0.384} & \multicolumn{1}{c|}{-} & \multicolumn{1}{c|}{-} \\ \cline{2-7} 
\multicolumn{1}{|l|}{} & \multicolumn{2}{c|}{T-test} & \multicolumn{1}{l|}{2.950***} & \multicolumn{1}{l|}{4.578***} & \multicolumn{1}{c|}{-} & \multicolumn{1}{c|}{-} \\ \hline
\multicolumn{7}{l}{Note: p\textless{}0.01: ***, 0.01\textless{}p\textless{}0.05: **}
\end{tabular}
\end{table*}

\begin{table*}[]
\centering
\caption{Logit regression results}

\begin{tabular}{llllllll}
\hline
\multicolumn{1}{|l|}{\textbf{dataset}} & \multicolumn{1}{l|}{\textbf{AUs}} & \multicolumn{1}{l|}{\textbf{AU 4}} & \multicolumn{1}{l|}{\textbf{AU 5}} & \multicolumn{1}{l|}{\textbf{AU 6}} & \multicolumn{1}{l|}{\textbf{AU 7}} & \multicolumn{1}{l|}{\textbf{AU 12}} & \multicolumn{1}{l|}{\textbf{AU 23}} \\ \hline
\multicolumn{1}{|l|}{\multirow{2}{*}{\textbf{JAFFE}}} & \multicolumn{1}{l|}{Coefficient} & \multicolumn{1}{l|}{-1.156***} & \multicolumn{1}{l|}{-0.415***} & \multicolumn{1}{l|}{0.084} & \multicolumn{1}{l|}{0.571***} & \multicolumn{1}{l|}{1.743***} & \multicolumn{1}{l|}{-0.450***} \\ \cline{2-8} 
\multicolumn{1}{|l|}{} & \multicolumn{1}{l|}{S.E.} & \multicolumn{1}{l|}{0.1} & \multicolumn{1}{l|}{0.039} & \multicolumn{1}{l|}{0.066} & \multicolumn{1}{l|}{0.038} & \multicolumn{1}{l|}{0.101} & \multicolumn{1}{l|}{0.055} \\ \hline
\multicolumn{1}{|l|}{\multirow{2}{*}{\textbf{DEFE}}} & \multicolumn{1}{l|}{Coefficient} & \multicolumn{1}{l|}{-1.6**} & \multicolumn{1}{l|}{0.373} & \multicolumn{1}{l|}{33.442} & \multicolumn{1}{l|}{31.959} & \multicolumn{1}{l|}{-15.207} & \multicolumn{1}{l|}{0.978} \\ \cline{2-8} 
\multicolumn{1}{|l|}{} & \multicolumn{1}{l|}{S.E.} & \multicolumn{1}{l|}{0.631} & \multicolumn{1}{l|}{0.729} & \multicolumn{1}{l|}{3961.164} & \multicolumn{1}{l|}{3826.095} & \multicolumn{1}{l|}{2037.702} & \multicolumn{1}{l|}{1.516} \\ \hline
\multicolumn{8}{l}{Note: p\textless{}0.01: ***, 0.01\textless{}p\textless{}0.05: **}
\end{tabular}
\end{table*}

\subsection{Result and Discussion}

\begin{figure}[!t]

\centering
\includegraphics[width=0.3\textwidth]{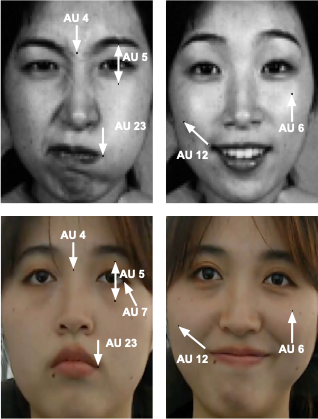}
\caption{Sample images of facial expressions in JAFFE (1st row) and DEFE (2nd row) with labelled AUs. Left: anger, right: happiness.}
\label{fig:label7}
\end{figure}

\par Statistical analysis results of AUs presence are shown in Table 6. For happiness, the results show that AU6 and AU12 movements could be observed in both JAFFE and DEFE. However, compared with JAFFE, the presence frequencies of AU 6 and AU 12 in DEFE were significantly lower (p\textless0.01). For anger, the results show that AU4, AU5 and AU23 movements could be observed in both JAFFE and DEFE, and there are significant differences (p\textless0.01). Besides, we found that AU7 related to anger from DEFE did not appear in the anger expressions from JAFFE. Sample images of facial expressions in JAFFE and DEFE with labelled AUs as shown in Figure.\ref{fig:label7}.

\par Compared with JAFFE, DEFE had a lower presence frequency on AU4, AU5, AU6, and AU12, especially AU4, which is highly related to anger, had a slight presence frequency in DEFE. The results may be caused by the main driving task, which requires concentration during driving, and the concentration may decrease the presence of AUs near eyes. On the other hand, the presence frequencies of AU7 and AU23 were lower in JAFFE, which maybe because of the difficulties to express negative emotions in Japanese culture \cite{matsumoto1989american}.

\par The logit regression results are shown in Table 7. According to our regression results, in JAFFE, for happiness, the coefficients of AU6 and AU12 were consistent with the results from FACS \cite{barrett2019emotional}, which means AU6 and AU12 were related to happiness. However, only the results of AU12 are significant (p\textless0.01). For anger, the coefficients of AU4, AU5, and AU23 were consistent with the results from FACS \cite{barrett2019emotional}, which means AU4, AU5, and AU23 were related to happiness.  The results of AU4, AU5, and AU23 are significant (p\textless0.01). Interestingly, AU7 (lid tightener) presence shows that AU7 was related to happiness which was different from previous researches\cite{barrett2019emotional}. In DEFE, only the result of AU4 was significant(0.01\textless p\textless0.05), and the coefficient was consistent with the research in FACS, indicating that AU4 had a significant predictive ability for anger. Other AUs were not observed with significant results.

\par Overall, for AUs presence, AU4 (Brow Lowerer), AU5 (Upper Lid Raiser), AU6 (Cheek Raiser), AU7 (Lid Tightener), AU12 (Lip Corner Puller), AU23 (Lip Tightener) are significant differences between dynamic driving and static life scenarios. The presence of AU4, AU5, AU6 and AU12 are higher in static life scenarios, indicating that AU AU4, AU5, AU6 and AU12 in dynamic driving scenarios are affected by the main driving tasks, which suppresses the facial expression of the driver ’s emotions. Meanwhile, the presence of AU7 and AU23 is higher in dynamic driving scenarios, which may be because Japanese culture suppresses the expression of negative emotions \cite{matsumoto1989american}. As for logit regression results, there are also significant differences between dynamic driving and static life scenarios. For anger, the results in dynamic driving scenarios show that only AU4 is significantly related to anger, while in static life scenarios AU4, AU5, and AU23 are all significantly related to anger. For happiness, the logistic regression results in dynamic driving scenarios show that there is no significant correlation between AUs and happiness, but the results in static life scenarios show that AU12 is significantly related to happiness. These significant differences were most likely due to the main driving tasks, which reduced the frequency and amplitude of facial muscle movements. Due to the limitation of JAFFE data amount, these results may require further investigations.

\section{Conclusion and future work}
In this work, a dataset for the analysis of spontaneous driver emotions elicited by video-audio stimuli is presented. The dataset includes facial expression recordings from 60 participants during driving. After watching each of the three video-audio clips selected to elicit specific emotions, each participant completed the driving tasks in the same driving scenarios and rated their emotional response in this driving process from the aspects of dimensional emotion and discrete emotion. These self-reported emotions include the scales of arousal, valence, and dominance as well as emotion category and intensity. We selected these three video-audio chips using the SAM and DES scales, which ensured the effectiveness of these stimulus materials aimed at the Chinese cultural background. Besides, we conducted the classification experiment for the scales of arousal, valence, and dominance as well as emotion category and intensity to establish baseline results for the proposed dataset in terms of accuracy and F1 scores, and these results were significantly higher than the results for random classification.

\par Moreover, we also compared the classification results in terms of accuracy and F1 score of the DEFE dataset with the DEAP and CK+ datasets, and the results show that the recognition results of the DEFE dataset are lower than the CK + dataset. Furthermore, we discussed the differences in facial expressions between driving and non-driving scenarios by comparing the presence of AU in the DEFE and JAFFE datasets. The results show that there were significant differences in AUs presence of facial expressions between driving and non-driving scenarios, and the difference will affect the results of facial emotion prediction, indicating that human emotional expressions in driving scenarios were different from other life scenarios. Therefore, publishing a human emotion dataset specifically for the driver is necessary for traffic safety improvement.

\par The DEFE dataset will be made publicly available after the work is published to allow researchers to evaluate their algorithms on an off-the-shelf driver facial expression dataset and investigate the possibility of applying them to applications. The DEFE data set provides the possibility to study emotion recognition from different emotion models simultaneously. Meantime, DEFE data can also be used to analyze the difference between driving and non-driving. Also, there are facial occlusions in DEFE, such as glasses and hands, which increases the complexity of facial expression recognition which is a significant research challenge.

\section*{Acknowledgment}

The authors would like to thank Peizhi Wang, Qianjing Hu, Mingqing Tang, Bingbing Zhang, Guanzhong Zeng and Mengna Liao for their assistance.

\ifCLASSOPTIONcaptionsoff
  \newpage
\fi



%

\bibliography{DEFE_dataset.bib}
\bibliographystyle{ieeetr}




%

\begin{IEEEbiography}[{\includegraphics[width=1in,height=1.25in,clip,keepaspectratio]{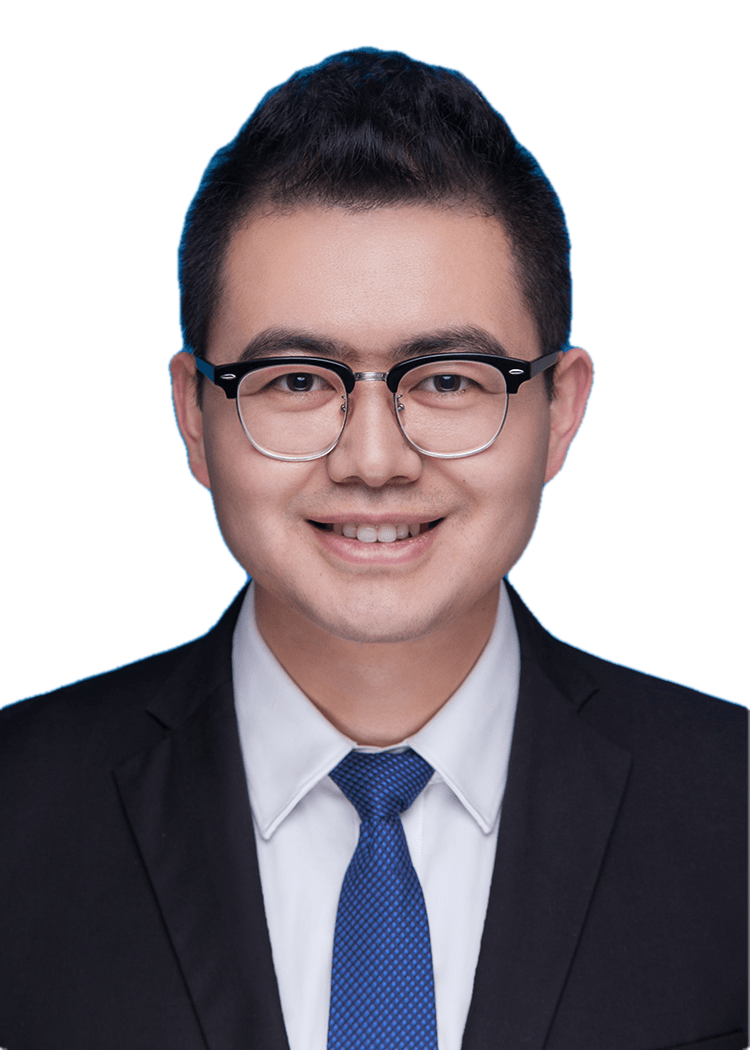}}]{Wenbo Li}
received the B.S. and M.Sc. degree in automotive engineering from Chongqing University, Chongqing, China, in 2014, and 2017, respectively. He is currently working toward the Ph.D. degree with the Advanced Manufacturing and Information Technology Laboratory,Department of Automotive Engineering, Chongqing University, Chongqing, China. He is also a visiting Ph.D. student at the Waterloo Cognitive Autonomous Driving (CogDrive) Lab at University of Waterloo, Canada. His research interests include intelligent vehicle, human emotion, driver emotion detection, emotion regulation, human-machine interaction, brain computer interface.
\end{IEEEbiography}

\begin{IEEEbiography}[{\includegraphics[width=1in,height=1.25in,clip,keepaspectratio]{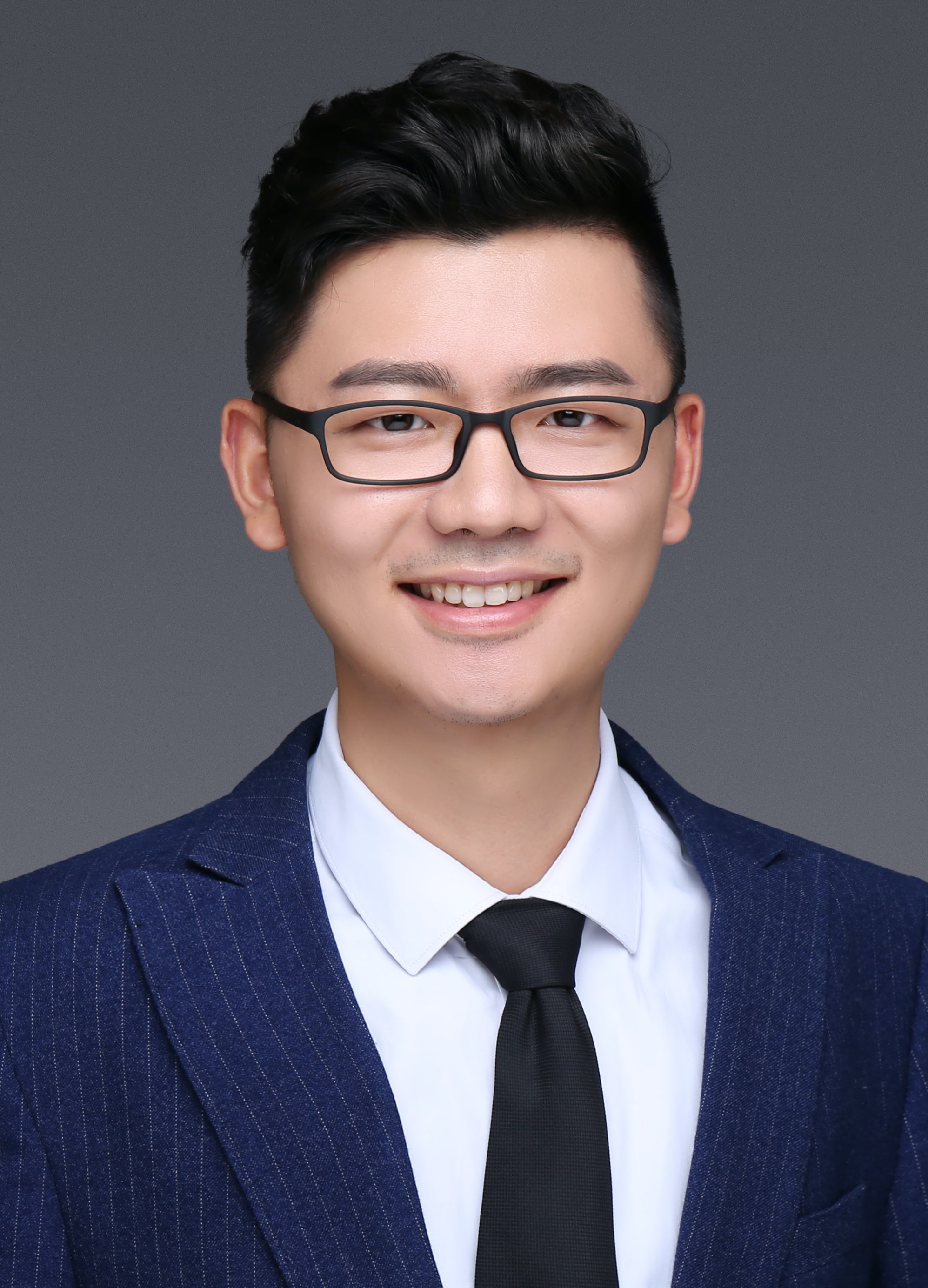}}]{Yaodong Cui}
received the B.S. degree in automation from Chang'an University, Xi'an, China, in 2017, received the M.Sc. degree in  Systems, Control and Signal Processing from the University of Southampton, Southampton, UK, in 2019. He is currently working toward a Ph.D. degree with the Waterloo Cognitive Autonomous Driving (CogDrive) Lab, Department of Mechanical Engineering, University of Waterloo, Waterloo, Canada. His research interests include sensor fusion, perception for the intelligent vehicle, driver emotion detection.
\end{IEEEbiography}

\begin{IEEEbiography}[{\includegraphics[width=1in,height=1.25in,clip,keepaspectratio]{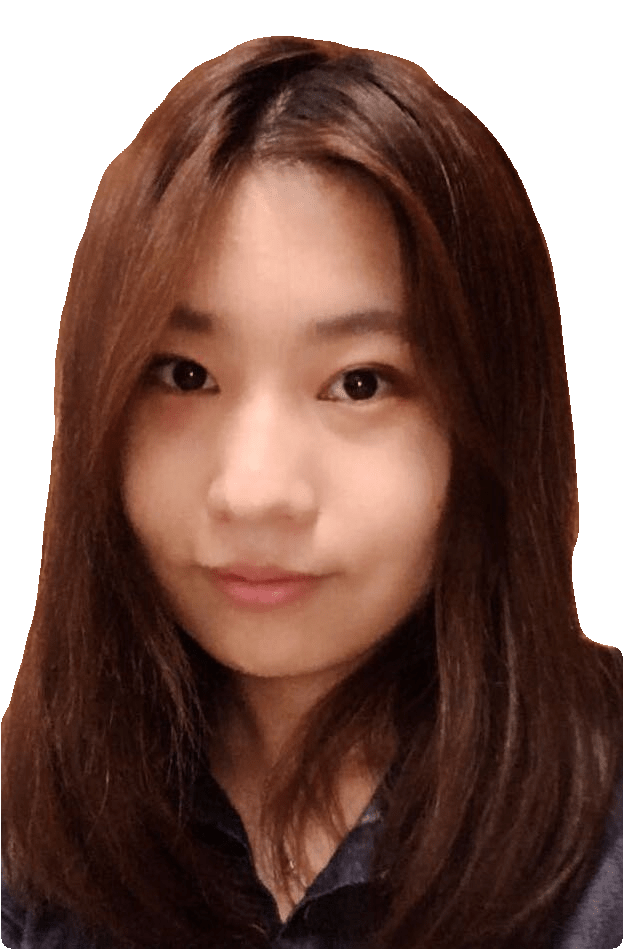}}]{Yintao Ma}
received the B.Sc. degree in Engineering Mechanics from the University of Illinois in Urbana Champaign, USA, in 2018. She is currently working toward the M.Sc. degree with the Cognitive Autonomous Driving Laboratory, Department of Mechanical and Mechatronics Engineering, University of Waterloo, ON, Canada. Her research interests include machine learning, image processing, and facial expression recognition.
\end{IEEEbiography}

\begin{IEEEbiography}[{\includegraphics[width=1in,height=1.25in,clip,keepaspectratio]{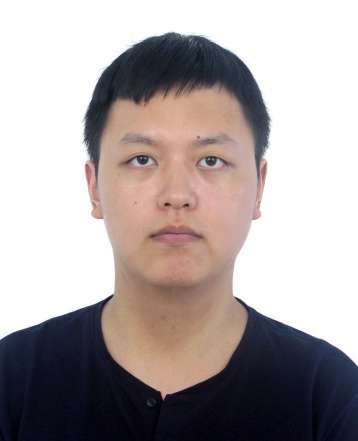}}]{Xingxin Chen}
received the B.Sc. degree from Nanjing University, Nanjing, China, in 2018. He is a Master of Applied Science (MASc) student in Waterloo Cognitive Autonomous Driving (CogDrive) Laboratory, Department of Mechanical and Mechatronics Engineering, University of Waterloo, Canada. His research interests include domain adaptation, transfer learning, computer vision.
\end{IEEEbiography}

\begin{IEEEbiography}[{\includegraphics[width=1in,height=1.25in,clip,keepaspectratio]{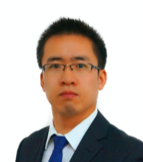}}]{Guofa Li}
(M’18) received the Ph.D. degree in Mechanical Engineering from Tsinghua University, Beijing, China, in 2016. He is currently an Assistant Professor in mechanical engineering and automation with the College of Mechatronics and Control Engineering, Shenzhen University, Guangdong, China. His research interests include driving safety in autonomous vehicles, driver behavior and decision making, computer vision, machine learning, and human factors in automotive and transportation engineering. He is the recipient of the Young Elite Scientists Sponsorship Program by SAE-China (2018), the Excellent Young Engineer Innovation Award from SAE-China (2017), and the NSK Sino-Japan Outstanding Paper Prize from NSK Ltd. (2014).
\end{IEEEbiography}

\begin{IEEEbiography}[{\includegraphics[width=1in,height=1.25in,clip,keepaspectratio]{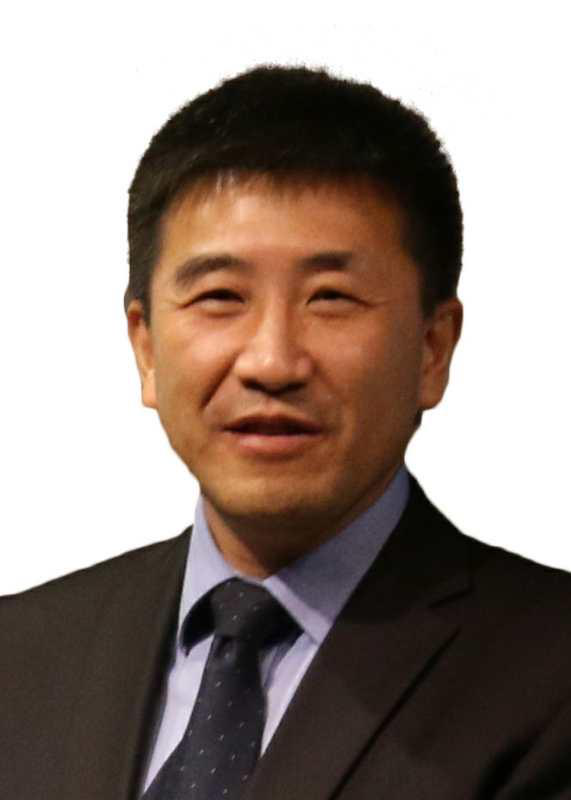}}]{Gang Guo}
received the B.S., M.S., and Ph.D degrees in automotive engineering from Chongqing University, Chongqing, China, in 1982, 1984, and 1994, respectively. He is currently the Chair and professor at the Department of Automotive Engineering, Chongqing University. He also serves as the Associate Director for the Chongqing Automotive Collaborative Innovation Center. He has authored and co-authored over 100 refereed journal and conference publications. His research interests include intelligent vehicle, multi-sense perception, human-machine interaction, brain computer interface, intelligent manufacturing, and user experience. Dr. Guo is a senior member of the China Mechanical Engineering Society and the Director of the China Automotive Engineering Society. He is also a member of the China User Experience Alliance Committee.
\end{IEEEbiography}

\begin{IEEEbiography}[{\includegraphics[width=1in,height=1.25in,clip,keepaspectratio]{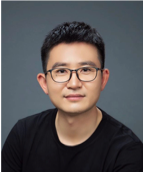}}]{Dongpu Cao}
(M’08) received the Ph.D. degree from Concordia University, Canada, in 2008. He is the Canada Research Chair in Driver Cognition and Automated Driving, and currently an Associate Professor and Director of Waterloo Cognitive Autonomous Driving (CogDrive) Lab at University of Waterloo, Canada. His current research focuses on driver cognition, automated driving and cognitive autonomous driving. He has contributed more than 200 papers and 3 books. He received the SAE Arch T. Colwell Merit Award in 2012, and three Best Paper Awards from the ASME and IEEE conferences. Dr. Cao serves as an Associate Editor for IEEE TRANSACTIONS ON VEHICULAR TECHNOLOGY, IEEE TRANSACTIONS ON INTELLIGENT TRANSPORTATION SYSTEMS, IEEE/ASME TRANSACTIONS ON MECHATRONICS, IEEE TRANSACTIONS ON INDUSTRIAL ELECTRONICS, IEEE/CAA JOURNAL OF AUTOMATICA SINICA and ASME JOURNAL OF DYNAMIC SYSTEMS, MEASUREMENT AND CONTROL. He was a Guest Editor for VEHICLE SYSTEM DYNAMICS and IEEE TRANSACTIONS ON SMC: SYSTEMS. He serves on the SAE Vehicle Dynamics Standards Committee and acts as the Co-Chair of IEEE ITSS Technical Committee on Cooperative Driving.
\end{IEEEbiography}





\end{document}